%% file: main.tex
\documentclass[conference]{IEEEtran}
\IEEEoverridecommandlockouts
\usepackage{cite}
\usepackage{amsmath,amssymb,amsfonts}
\usepackage{algorithmic}
\usepackage{graphicx}
\usepackage{textcomp}
\usepackage{xcolor}
\usepackage{dsfont}
\def\BibTeX{{\rm B\kern-.05em{\sc i\kern-.025em b}\kern-.08em
    T\kern-.1667em\lower.7ex\hbox{E}\kern-.125emX}}
\begin{document}

\title{Generalization and Membership Inference Attack a Practical Perspective \\
}

\author{
\IEEEauthorblockN{Fateme Rahmani}
\IEEEauthorblockA{\textit{Department of Computer Engineering} \\
\textit{Sharif University of Technology}\\
Tehran, Iran \\
f.rahmani@sharif.edu}
\and
\IEEEauthorblockN{Mahdi Jafari Siavoshani}
\IEEEauthorblockA{\textit{Department of Computer Engineering} \\
\textit{Sharif University of Technology}\\
Tehran, Iran \\
mjafari@sharif.edu}
\and
\IEEEauthorblockN{Mohammad Hossein Rohban}
\IEEEauthorblockA{\textit{Department of Computer Engineering} \\
\textit{Sharif University of Technology}\\
Tehran, Iran \\
rohban@sharif.edu}
}

\maketitle

\begin{abstract}
With the emergence of new evaluation metrics and attack methodologies for Membership Inference Attacks (MIA), it becomes essential to reevaluate previously accepted assumptions.
In this paper, we revisit the longstanding debate regarding the correlation between MIA success rates and model generalization using an empirical approach. We focused on employing augmentation techniques and early stopping to enhance model generalization and examined their impact on MIA success rates. 
We found that utilizing advanced generalization techniques can significantly decrease attack performance, potentially by up to 100 times. Moreover, combining these methods not only improves model generalization but also reduces attack effectiveness by introducing randomness during training. Additionally, our study confirmed the direct impact of generalization on MIA performance through an analysis of over 1K models in a controlled environment.

\end{abstract}

\begin{IEEEkeywords}
Membership inference attack, Privacy, Machine learning, Generalization
\end{IEEEkeywords}

\section{Introduction}

As machine learning models become more common in different applications, there is growing attention to their security. One notable security concern is the susceptibility of ML models to membership inference attacks (MIA).
MIA exploits information leaks from models to infer if the specific data points were part of the training dataset, posing a threat to individuals' privacy.
Despite their simplicity, MIAs play a fundamental role in more severe security breaches, like training data extraction \cite{carlini2019secret, carlini2021extracting}.

In most research on membership inference attacks, the categorization is based on the varying degrees of attackers' access and capabilities. 
Attacks are categorized based on the type of attacker access level \cite{nasr2019comprehensive,  sablayrolles2019white, song2021systematic, leino2020stolen} (black-box vs white-box), the nature of the attack  \cite{nasr2019comprehensive, pustozerova2020information, chen2020beyond} (passive or active), and the timing of the attack \cite{lira2022, ye2022enhanced} (online or offline).
However, we provided a novel classification based on the attack approach considering the latest attacks introduced recently. These approaches might be applicable to different access levels and timings.

Lira \cite{lira2022}, one of the most recent studies in this field, introduced a new metric for assessing MIA, emphasizing the privacy implications of attacks. With a focus on privacy, having minimal information about the entire population is generally not considered a threat. However, obtaining specific details about a small segment of society is a privacy violation. Therefore, as commonly done in other machine learning studies, relying on attack accuracy may not be the optimal choice, given that balanced accuracy is an average-case metric with symmetrical characteristics \cite{lira2022}. Lira further introduced a new metric for evaluating attacks, focusing on confident attack results where the false positive rate is notably low.
They also introduced a new attack that integrates established attack methods with the concept of per-sample hardness, delivering a robust attack surpassing all current methods. Despite its simplicity, according to our assessments, this black-box attack can outperform all current white-box MI attacks.

With this new metric, we should reexamine all our earlier assumptions and findings. As per Lira's review, specific attacks previously thought to be weak turn out to be stronger than we expected.

This paper empirically investigates the well-known debate about the correlation between generalization and the MIA success rate.
Utilizing the state-of-the-art Lira attack as the MIA approach, we assess the attack's effectiveness using the newly introduced metric TPR @ 0.1\% FPR.
Our focus in this study is on the impact of augmentation techniques and early stopping as methods to enhance model generalization. Finally, we explore the relationship between generalization and MIA success rates by attacking over 1K models trained with different methods (while controlling other contributing factors, such as model architecture and training samples).

In the following sections, we first review existing MIA approaches, categorizing them based on their attack perspective. We then survey evaluation metrics and their significance in assessing the MIA. Next, we revisit the ongoing debate in past studies regarding the connection between MIA and model generalization. The following section summarizes our experimental findings in three key steps. Initially, we investigate how augmentation and early stopping impact attack performance. Subsequently, we explore the influence of complex augmentation techniques on the overall attack efficacy. Lastly, we examine the correlation between attack success rate, generalization, and test accuracy.

\input{litrature}

\input{experiments}

\section{Conclusion}

In conclusion, our study reveals that we can simultaneously improve model performance and safeguard privacy by adopting simple methods that enhance model generalization. Using state-of-the-art augmentation techniques can reduce the attack effectiveness up to 100 times. Moreover, our experiments demonstrate that even with knowledge of the target model's training procedure and utilizing complex training techniques, the attacker cannot enhance the MIA. This highlights the importance of employing combinations of simple methods to reduce the attack effectiveness by introducing randomness to model training.

Although several factors influence the success of membership inference attacks, our controlled experimental setup shows that model generalization remains a key determinant of attack effectiveness. In particular, we observe that reducing the generalization gap consistently lowers the success rate of these attacks. Therefore, maintaining a small gap between training and test performance is a simple and effective way to mitigate privacy risks.

In real-world scenarios, we are often required to train machine learning models on fixed datasets, regardless of the potential privacy vulnerabilities of individual data points. In this work, we adopt this realistic setting by fixing the training dataset and evaluating attack performance under controlled conditions. Our results demonstrate that, even when the data contains vulnerable samples, effective training strategies that improve generalization can significantly reduce the risk of membership inference attacks.



\bibliographystyle{IEEEtran}
\bibliography{references}

\end{document}

%% file: litrature.tex
\section{Membership Inference Attack}

\begin{table*}
\renewcommand{\arraystretch}{1.7}
\centering
\caption{Comparison of Attack Categories}
\begin{tabular}{|p{2.5cm}|p{5cm}|p{4cm}|p{4cm}|}
\hline
\multicolumn{1}{|c|}{\textbf{Category}} & \multicolumn{1}{c|}{\textbf{Key Idea}} & \multicolumn{1}{c|}{\textbf{Advantage}} & \multicolumn{1}{c|}{\textbf{Limitations}} \\
\hline
NN-Based & Using the power of ML models against them, training a ML attack model & Easily utilizing multiple sources for inferring membership & High computational and time cost \\
\hline
Threshold-Based & Identifiable patterns in some simple metrics and defining fixed threshold based on them & Simplicity, ease of application & Challenge in threshold selection and adaptation \\
\hline
Adaptive Threshold-Based & Threshold adjusted based on inference difficulty & Improved performance by considering data complexities & More complex to implement, hardness estimation required \\
\hline
LRT-Based & Compare likelihood under membership and non-membership hypotheses & Considering samples hardness based on attacker information & High computational cost for training multiple shadow models \\
\hline
Vulnerability-Based & Targeting most vulnerable data points by estimating the hardness & Prioritize vulnerable points, resource conservation & Varying attack accuracy, incomplete assessment \\
\hline
\end{tabular}
\end{table*}

Membership Inference Attacks are a group of attacks aimed at determining whether a specific data point belongs to the training dataset of an ML model. These attacks can be categorized into several subtypes, each with its own approach and characteristics. We categorize MIAs into five categories based on their attack approach.


\subsubsection{NN-based Attacks}
Neural network-based attacks involve training an attack model using data from the shadow model. The attacker trains shadow models to mimic the target model's behavior.

The first NN-based attack, introduced by Shokri et al. \cite{shokri2017membership}, is based on training a set of shadow models and then comparing their behavior on members and non-members.
This attack primarily relies on the difference in confidence of the model's predictions. It expects the shadow models to behave similarly to the target model.
Salem et al. \cite{salem2018ml} showed that it is possible to achieve similar accuracy to \cite{shokri2017membership} using only a shadow model.

Nasr et al. \cite{nasr2019comprehensive} designed a white-box membership inference attack that utilizes target model parameters for inference, with separate components for each layer outputs and activations.
However, despite its broad access to target model parameters, they could not significantly improve the attack performance.

\subsubsection{Threshold-Based Attacks}
In these attacks, a threshold is set using a specific criterion for the target model. When the data score exceeds this threshold, the data is predicted to be a member. Various scores can be employed for this purpose including correctness \cite{yeom2020overfitting}, confidence \cite{song2019privacy, song2021systematic, yeom2018privacy, watson2021importance}, entropy \cite{shokri2017membership, song2021systematic}, and modified entropy \cite{song2021systematic}. Correctness is based on the model's output label, confidence measures the likelihood of the correct label, entropy quantifies prediction uncertainty, and modified entropy combines prediction entropy and true labels. 

The choice of threshold for each criterion significantly impacts the attack's success. Therefore, various methods are employed to select thresholds more effectively, including using multiple shadow models, applying different perturbations to the data \cite{jayaraman2020revisiting}, or optimizing thresholds independently for each class \cite{song2021systematic}.
In general, this attack can be summarized in this simple term:
\begin{equation}
	I(x,y) = \mathds{1}[\mathcal{A}(x,y) > \tau]
\end{equation}

The symbol $\mathcal{A}$ stands for any of the criteria mentioned, and $\tau$ represents the calculated threshold value based on that criterion.

\subsubsection{Adaptive Threshold-Based Attacks}

A fixed threshold was selected for all the data points examined in threshold-based attacks. However, it is essential to note that determining membership for specific points may be more challenging than others. Since some data points are harder to learn and classify, adjusting our approach to consider the difficulty of each sample can significantly enhance attack performance.

The idea of an attack relying on per-sample hardness was initially presented by Sablayrolles et al. \cite{sablayrolles2019white} 
They proposed an online attack strategy employing a calibrated score, where $\gamma(\cdot)$ computed for each point modifies to adjust the attack based on the sample's hardness.

\begin{equation}
	\mathcal{A}_{cal}(x,y) = \mathcal{A}(x,y) + \gamma(x,y)
\end{equation}

Furthermore, Watson et al. \cite{watson2021importance} introduced an offline attack considering the per-sample hardness. They assess how hard it is to attack a data point by estimating its expected score on shadow models.

\subsubsection{LRT-Based Attacks}
The central idea of these attacks is to create a test that distinguishes between two hypotheses: that a data point was sampled from the training dataset (H0) and the hypothesis that it was not (H1).

Ye et al. \cite{ye2022enhanced} converts the Likelihood Ratio Test into a threshold-based attack, where the threshold varies based on different information available to the attacker. This approach can guarantee a TPR at a fixed FPR. The threshold calculation involves training a specific number of shadow models to target a desired FPR.

In another method proposed by Lira \cite{lira2022}, they define distributions representing models that either have or have not been trained on the particular data of interest. These distributions help to convert the attack into a Likelihood Ratio Test. However, the defined test is unsolvable, and they simplified the test with new distributions based on the loss of models trained with and without the specific data.

\subsubsection{Vulnerability-Based Attacks}
In privacy-preserving scenarios, it is crucial to accurately infer the membership of data points, even when the number of such data points is limited. Long et al. \cite{long2020pragmatic} introduced an approach to attack by identifying data points that are more vulnerable and focusing the attack on these specific points. To detect data points with higher vulnerability, they construct a graph for target data points by cosine similarity of their features.

The method calculates the probability of membership of vulnerable data points with high confidence but refrains from making strong assumptions about the remaining data points. This approach may result in a lower average attack accuracy, but it is a strong attack with newer metrics.


\subsection{MIA Evaluation Metrics}

Various research studies have introduced different strategies for assessing the success of membership inference attacks. It is essential to note that in MIA, the goal is to identify members of the training dataset, while identifying non-members is irrelevant. Several evaluation strategies are commonly employed:

\begin{itemize}
	\item \textbf{Accuracy:} \cite{nasr2019comprehensive, sablayrolles2019white, song2021systematic, watson2021importance, choquette2021label}
   The simplest method for evaluating the effectiveness of a membership inference attack is accuracy, which measures the number of correct membership predictions in a balanced dataset of members and non-members. Since accuracy is a symmetric metric and assigns equal importance to FN and TN, it may not be the most appropriate metric for detecting members.
	
	\item \textbf{Precision and Recall:} \cite{shokri2017membership, salem2018ml, melis2019exploiting, yeom2018privacy, jayaraman2020revisiting, long2020pragmatic}
   Another method for evaluating the attack is based on two parameters: precision and recall. Precision indicates what percentage of the member-predicted data are members, while recall assesses how many members have been identified. Per the definitions, precision is a suitable metric for evaluating MIA, while recall is not relevant. In membership inference attacks, high precision achieved even with low recall indicates that privacy can be compromised on a small subset of the data.
	
	\item \textbf{ROC Curve:} \cite{ye2022enhanced, watson2021importance}
   Analyzing the ROC curve, which plots the TPR against the FPR for different thresholds on the attack score, provides valuable insights into evaluating membership inference attacks. Since predicting FP is of greater concern in these attacks, we are expected to evaluate the attack where the FPR is low enough to trust the attack results. A log-scaled ROC curve can be utilized to emphasize the low range of FPR.

  The Area under the ROC curve (AUC) provides a numerical measure for attack comparison. However, since the AUC is an average of all FPRs, it does not specifically focus on cases with low FPRs.
	
	\item \textbf{TPR at Low FPR:}
   Carlini et al. \cite{lira2022} have proposed a metric that examines the TPR at very low FPRs, such as $0.1\%$. This metric uses exact points on the ROC curve with low FPR to assess the performance of attacks more rigorously.
\end{itemize}

In conclusion, the best method for analyzing attack performance is to examine the ROC curve on a logarithmic scale. For numerical comparisons of attacks, precision with a fixed recall and the TPR at low fixed FPR, as introduced by Carlini et al. \cite{lira2022}, represent the most effective evaluation approaches. However, it is essential to note that, in most cases, finding the precision of a model on a fixed recall is not feasible; thus, assessing the TPR at low FPR is the most suitable numerical metric for comparing attacks. Following the Lira, TPR @0.1\% FPR will be utilized for evaluating attacks in subsequent sections.


\section{MIA \& Generalization}
Generalization in a machine learning model occurs when it can perform similarly on previously unseen data. When a model overfits the training data, it preserves unnecessary details of the training data rather than following the overall data pattern. Preserving training data leads to increased vulnerability to membership inference attacks, a concept explored in various prior research studies \cite{shokri2017membership, salem2018ml, yeom2018privacy, choquette2021label, lira2022}.

In \cite{yeom2018privacy}, the mathematical relationship between generalization and the mean error of generalization is explored concerning membership inference attacks. The study concludes that model generalization is a robust predictor of vulnerability to membership inference attacks.

Shokri et al. \cite{shokri2017membership} investigated the relationship between overfitting and their attack, highlighting that overfitting contributes to increased attack accuracy. However, overfitting alone is not the only factor influencing the attack. The structure and type of the target model also affect the attack accuracy. Two different models with similar overfitting levels may yield different attack accuracies, indicating that generalization is essential for mitigating the severity of membership inference attacks.

Research studies like \cite{salem2018ml} and \cite{choquette2021label} have examined similar aspects of overfitting and confirmed the claims made by Shokri et al. Furthermore, several other studies related to membership inference attacks have mentioned generalization as a method to reduce attack intensity.

However, a different perspective is introduced by Carlini et al. \cite{lira2022}. Examining various sample instances demonstrates that models with similar generalizations have vulnerabilities up to 100 times greater against MIA. Moreover, their analysis indicates that more accurate models are more susceptible to MIA.

The literature on membership inference attacks provides conflicting insights into the role of generalization, and it is clear that generalization is a critical aspect affecting the privacy-performance trade-off in machine learning models.


%% file: experiments.tex
\section{Experiments}

Various research studies evaluate privacy in ML algorithms using privacy-sensitive datasets. For instance, two datasets commonly employed are the Purchase Dataset (the purchase history of various individuals) and
Texas Hospital Stays Dataset (information about patients hospitalized in Texas hospitals from 2006 to 2009).
However, due to the simplicity of their structures, these datasets lack the necessary complexities for privacy investigation \cite{lira2022}.

In addition to these datasets, the CIFAR dataset has been utilized in previous research. CIFAR-10 is a dataset designed for image classification, consisting of 60k images with dimensions 32 $\times$ 32, categorized into 10 different classes. Out of these 60k, 50k are used for training and 10k for testing. CIFAR-100 follows a similar format, with 100 classes containing 600 images. In this research, we employ CIFAR to train target models.

The model must train on a subset of the data to evaluate MIA on the target model. As a larger dataset reduces the likelihood of overfitting, we use the entire dataset, including both training and testing data. Thus, we use half of the combined training and testing data (30k samples) to train each target model.

In this study, we utilize ResNet \cite{he2016deep} as the primary architecture for training our target models. ResNet, a deep neural network, has been chosen for its capability to overcome challenges associated with training deep models by integrating shortcuts. These shortcuts enable efficient gradient flow, ensuring that updates reach the initial layers, thereby facilitating the training of deeper neural networks.

The ResNet architecture is characterized by multiple blocks, each consisting of two convolutional layers and a batch normalization layer. Specifically, we adopt ResNet-18, which comprises 8 blocks, resulting in 18 layers.

\subsection{Attack Evaluation under Varying Conditions}

Previous research has often not explicitly detailed the precise state of the target model in reported results. The accuracy of the target model in terms of training and testing data significantly influences the final outcome of an attack. 
This variability makes it challenging to compare methodologies in the future, and it needs to be clarified how effectively the target model generalizes. To address this, we evaluate attacks on different target models in various conditions, reporting models training status and attack performance in each case. Throughout these evaluations, the architectures and training datasets of the target models remain consistent to ensure comparability.

\subsubsection{Augmentation Techniques}

As the quantity of training data increases, the risk of model overfitting diminishes. Utilizing data augmentation techniques not only enhances the model's training but can also provide some level of defense against MIA. Each technique has a different impact on model generalization and, consequently, on the results of MIA. To investigate this, we employ various augmentation techniques, including geometric transformations, random erasing (cutout), and newer methods such as AutoAugment \cite{cubuk2019autoaugment}, designed specifically for the CIFAR dataset (CIFAR-policy).
Furthermore, we explore the combined effects of these techniques, which result in more sophisticated augmentation strategies.

\begin{table}[htbp]
    \centering
    \caption{
        Different data augmentation techniques and the attack TPR @0.1\% FPR.
    }
    \label{augment_res_table}
    \begin{tabular}{|c|c|c|c|}
        \hline
        Technique & Train Acc & Test Acc & TPR@FPR \\
        \hline
        \hline
        No Augmentation
        & 100 & 76.55 & 19.56 \\
        Crop
        & 95.75 & 84.26 & 0.73 \\
        Mirror
        & 98.75 & 72.59 & 4.52 \\
        CIFAR-Policy
        & 99.4 & 80.75 & 2.91 \\
        Cutout
        & 99.2 & 76.49 & 9.95 \\
        Crop + Cutout
        & 96.81 & 85.29 & 0.63 \\
        Crop + Mirror
        & 92.93 & 82.43 & 0.33 \\
        Cutout + Mirror
        & 98.9 & 81.43 & 4.61 \\
        Crop + Cutout + Mirror
        & 95.35 & 84.16 & 0.54 \\
        All Techniques
        & 94.94 & 89.15 & 0.18 \\
        \hline
    \end{tabular}
\end{table}

\begin{figure}
	\centering
	\includegraphics[width=8cm]{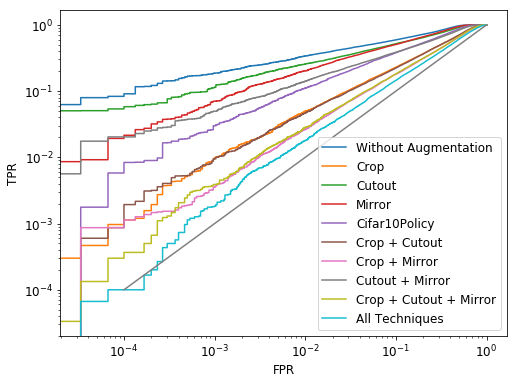}
	\caption{
		ROC curves in logarithmic scale for MIA on target models trained with different data augmentation techniques.
	}
	\label{augment_res_log}	 
\end{figure}

As observed in Table \ref{augment_res_table} and Figure \ref{augment_res_log}, proper usage of data augmentation techniques can reduce the accuracy of MIA by up to 100 times. Additionally, the same attack reported by Lira \cite{lira2022} has reached 8.4 TPR @0.1\% FPR, but this value is reducible to 0.18 with proper training.

\subsubsection{Early Stopping}

\begin{figure}[h]
	\centering
	\includegraphics[width=8cm]{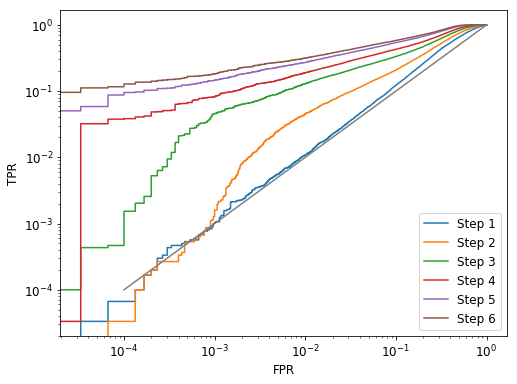}
    \caption{
		ROC curve of an MIA on a target model at different training steps; the train-test gap, test accuracy, and attack TPR @0.1\% FPR for each curve are specified in \ref{early_stopping_res_table}.
	}
	\label{training_steps}	 
\end{figure}

To assess the impact of early stopping on the performance of the MI attack, we stored a model in different training checkpoints and conducted attacks at each saved checkpoint. The results of the attacks at different training stages are illustrated in Figure \ref{training_steps}.

\begin{table}[htbp]
    \centering
    \caption{
        Early stopping and the attack TPR @0.1\%FPR.
    }
    \label{early_stopping_res_table}
    \begin{tabular}{|c|c|c|c|}
        \hline
        Step & Train Acc & Test Acc & T@F \\
        \hline
        \hline
        6
        & 100 & 76.8 & 18.28 \\
        5
        & 99.8 & 75.7 & 14.63 \\
        4
        & 97.4 & 73.2 & 8.31 \\
        3
        & 93.6 & 72.3 & 4.6 \\
        2
        & 86 & 68.3 & 0.16 \\
        1
        & 75.9 & 67.4 & 0.1 \\
        \hline
    \end{tabular}
\end{table}

As illustrated in Figure \ref{training_steps} and Table \ref{early_stopping_res_table}, it is evident that attacking a model with a higher train-test gap is more straightforward. Early stopping can effectively mitigate attacks by reducing the model's vulnerability to MIA.

\subsection{Attacker Knowledge of Training Technique}

Knowing how the target model is trained can help create more accurate shadow models, making the attack more precise. For instance, if the attacker knows the target model uses the CIFAR-Policy augmentation technique, they can use the same method for training shadow models. We expect that using similar settings will result in shadow models behaving similarly, and the resulting shadow models will be closer to the target model in model space.

\begin{figure}
	\centering
	\includegraphics[width=8cm]{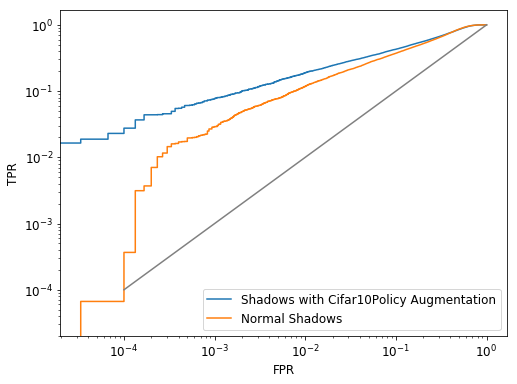}
	\caption{
		ROC curves of the attack on the target model trained with the CIFAR-Policy data augmentation technique in two scenarios: 1- Training shadow models conventionally, and 2- Training shadow models with the CIFAR-Policy data augmentation technique.
	}
	\label{cifar-augment}	 
\end{figure}

This expectation holds true for straightforward training techniques like using only one augmentation method, such as CIFAR-Policy. Figure \ref{cifar-augment} displays the difference in ROC curves when attacking a target model trained with the CIFAR-Policy data augmentation technique, both with and without this knowledge. Here, we can see that the attacker, armed with this knowledge, can boost the attack accuracy by up to 5 times.

\begin{figure}
	\centering
	\includegraphics[width=8cm]{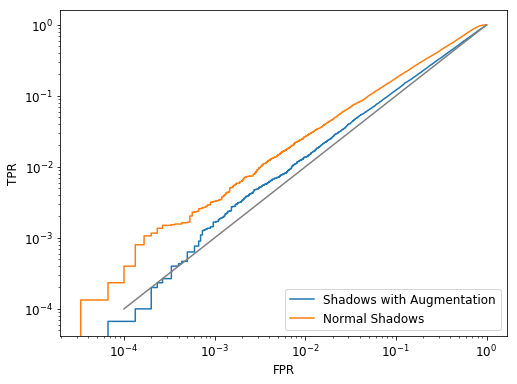}
	\caption{
		ROC curves of the attack on the target model trained with the combination of all data augmentation techniques in two scenarios: 1- Training shadow models conventionally, and 2- Training shadow models with a similar technique.
	}
	\label{all-augment}	 
\end{figure}

\begin{figure*}
	\centering
	\includegraphics[width=\textwidth]{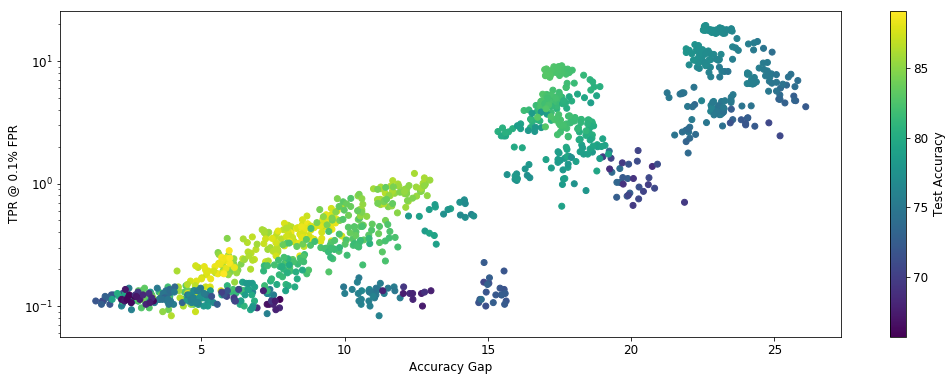}
	\caption{   
		Relationship of MIA TPR @0.1\% FPR and model accuracy gap for over 1K models. The model's test accuracy is color-coded in the plot. The vertical axis is on a logarithmic scale. The attack success rate might vary up to 10 times for models with the same accuracy gap.
	}
	\label{acc_gap}	 
\end{figure*}

However, this phenomenon is not observed in attacking models trained with more advanced data techniques. Figure \ref{all-augment} shows the difference in ROC curves for attacking a model trained with all mentioned data techniques, both with and without knowing those techniques. Clearly, when shadow models are trained similarly, the attack accuracy decreases. 

Since each augmentation technique involves a level of randomness, employing various techniques complicates the exact reproduction of data seen by the target model. This complexity likely contributed to the unexpected outcome observed in our study. Moreover, data augmentation can obscure the inherent difficulty of samples, reducing the attacker’s ability to exploit this signal; consequently, attackers may achieve better performance \textbf{without} applying augmentation.

In conclusion, we find that by combining data augmentation techniques and using up-to-date techniques that increase model generalization, we can simultaneously enhance model performance and privacy. The use of a combination of data augmentation techniques not only reduces the risk of overfitting but also, by making data regeneration more challenging, decreases the performance of the attack. \textbf{
Utilizing complex training techniques, the attacker, even with knowledge of the target model's training procedure, cannot enhance the MIA}.

\subsection{Generalization and MIA}

\begin{figure*}
	\centering
	\includegraphics[width=\textwidth]{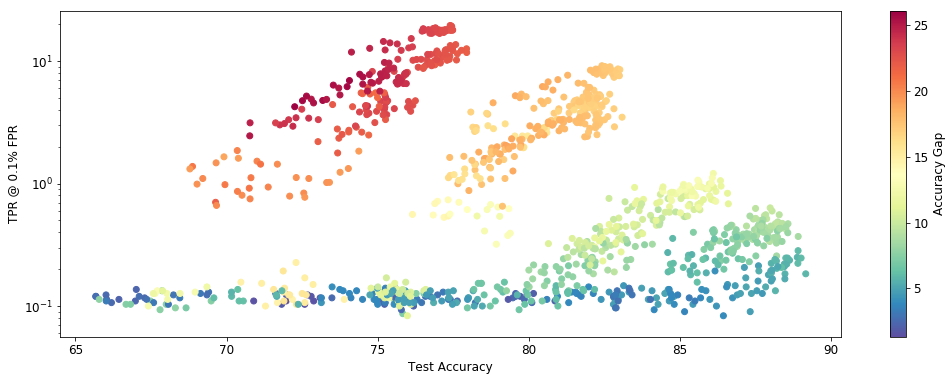}
	\caption{   
		Relationship of MIA TPR @0.1\% FPR and model test accuracy for over 1K models. The model's accuracy gap is color-coded in the plot. The vertical axis is on a logarithmic scale, the attack success rate might vary up to 100 times for models with same test accuracy.
	}
	\label{test_acc}	 
\end{figure*}

\begin{figure*}
	\centering
	\includegraphics[width=\textwidth]{images/loss_gap2.png}
	\caption{   
		Relationship of MIA TPR @0.1\% FPR and model loss gap for over 1K models. The model's test accuracy is color-coded in the plot. The vertical axis is on a logarithmic scale. The attack success rate might vary up to 10 times for models with the same loss gap.
	}
	\label{loss_gap}	 
\end{figure*}

In the previous section, we mentioned that employing state-of-the-art training techniques can reduce the success rates of MI attacks. These techniques notably enhance the generalizability of the target model, as indicated by train-test accuracies. However, further experiments are needed to investigate the relationship between generalization and MIA success rates. Since generalization is not the only factor influencing MIA performance, we have kept other contributing factors constant.

As highlighted in \cite{shokri2017membership}, the structure and capacity of the target model are additional influential factors impacting the accuracy of the attack. Moreover, the dataset itself and its size can significantly affect MIA performance. 
Moreover, as recent works have stated, the attack's success rate depends on the difficulty of the targeted samples. Therefore, we used the same training and testing set in all experiments to examine generalization effects more precisely.

To numerically evaluate the model's generalization, two metrics have been used by different studies: 1) the train-test accuracy gap \cite{shokri2017membership, leino2020stolen} and 2) the train-test loss gap \cite{yeom2018privacy, sablayrolles2019white}.

\begin{equation}
	Gap_{acc} = \underset{ (x,y) \in \mathcal{D}}{\mathbb{E}}[F(x)_y] - 
	\underset{ (x,y) \notin \mathcal{D}}{\mathbb{E}}[F(x)_y]
\end{equation}

\begin{equation}
	Gap_{loss} = \underset{ (x,y) \notin \mathcal{D}}{\mathbb{E}}\left[\ell\left(F(x), y\right)\right]-
	\underset{ (x,y) \in \mathcal{D}}{\mathbb{E}}\left[\ell\left(F(x), y\right)\right]
\end{equation}

These metrics can only partially illustrate the generalization as it is a more complicated concept, but they are an acceptable estimation of the model's overall generalization. 

We conducted the attack scenario on over 1K models with identical architectures and trained on similar trainsets. The results are illustrated in Figure \ref{acc_gap}, where we plotted the models' attack results based on the train-test accuracy gap. Each point is also color-coded to represent the model's test accuracy. In Figure \ref{acc_gap}, we observe a clear upward trend in the attack success rate as the model's generalization gap increases, indicating an increased vulnerability to MIA.

Furthermore, the attack success rate exhibits some dependence on test accuracy. As shown in \ref{acc_gap}, among models with the same accuracy gap, those with higher test accuracy tend to achieve higher attack performance. However, as illustrated in \ref{test_acc}, test accuracy alone is not a reliable predictor of a model’s vulnerability to MIA. In contrast, the color-coded accuracy gaps reveal a clear and consistent trend, indicating that the generalization gap remains the dominant factor influencing attack success.

Based on the reported results, in contrast to the findings in \cite{lira2022}, we conclude that \textbf{the accuracy gap, compared to the model's test accuracy, is a better predictor of the model's vulnerability to MIA}.
We believe the difference in our findings is rooted in test configuration differences. With the new metric introduced by Lira, the attack's success rate relies heavily on the hardness of the training samples. We suspect that the results in \cite{lira2022} are from models trained on different training and testing sets, which means they attacked different sets of points with different weaknesses.

A comparable plot, derived from the loss gap, is also depicted in Figure \ref{loss_gap}. As evident in both plots, the outcomes of the accuracy gap and loss gaps' outcomes closely align. However, the loss gap, which is more aligned with the definition of generalization, exhibits smoother results than the accuracy gap.  
Nonetheless, employing distinct loss functions can lead to a significant variation in the scale of the loss gap, making it more difficult or even impossible to compare different target models using this metric. Therefore, the accuracy gap serves as a reliable estimate of the model's susceptibility to MIA.
